\documentclass[letterpaper, 10 pt, conference]{ieeeconf}  

\IEEEoverridecommandlockouts                              

\usepackage{graphicx} 
\usepackage{amsmath}
\usepackage{amssymb}
\usepackage{bm}
\usepackage{bbm}
\usepackage{bbding} 
\usepackage{tikz}
\usepackage{siunitx} 
\usepackage{algorithm}
\usepackage{booktabs}
\usepackage{hyperref}
\usepackage[dvipsnames]{xcolor}
\usepackage{mathtools}
\usepackage{multirow}
\usepackage[noend]{algpseudocode}

\DeclareMathOperator*{\argmin}{arg\,min}

\newcommand{\blockagentwise}{First Block}

\newcommand{\blocktraj}{Sub-Block A}

\newcommand{\blocksafe}{Sub-Block B}

\newcommand{\blockglobal}{Second Block}

\title{\LARGE \bf
Sampling at Scale: Distributed Multi-Agent Trajectory Optimization

}

\author{Minchan Jung$^{1}$, Dimitrios S. Georgiou$^{1}$, Augustinos D. Saravanos$^{2}$ and Evangelos A. Theodorou$^{1}$ 
\thanks{$^{1}$Daniel Guggenheim School of Aerospace Engineering, Georgia Institute of Technology, Atlanta, GA, USA. {\tt\small \{minchan.jung, dgeorgiou3, evangelos.theodorou\}@gatech.edu}}%
\thanks{$^{2}$Department of Aeronautics and Astronautics, Massachusetts Institute of Technology, Cambridge, MA, USA. {\tt\small asaravan@mit.edu}}%
\thanks{Corresponding author: Minchan Jung}
}

\begin{document}

\maketitle
\thispagestyle{empty}
\pagestyle{empty}

\begin{abstract}
Multi-agent trajectory optimization poses two coupled challenges: effectively exploring non-convex landscapes and scaling favorably to large teams of agents.
Sampling-based methods are natural candidates for the exploration challenge but scale poorly with problem dimension. Moreover, their off-the-shelf integration into distributed optimization introduces a fundamental trade-off: 
sampling either fails to address the constraint-induced non-convexity or its search space grows with neighborhood size.
We address this gap by introducing a \underline{D}istributed \underline{S}ampling-based \underline{T}rajectory \underline{O}ptimization (DS-TO) architecture built on a splitting mechanism tailored to sampling.
Under this splitting, each agent samples only in its own control space, independent of neighborhood size, while still remaining exposed to both the constraint-induced non-convexity and the decisions of other agents.
The resulting algorithm is fully decentralized, requiring only neighborhood communication, with all local updates executed in parallel across agents. 
Efficient variations, including parallel Jacobi updates within each agent's subproblem and single-step constraint linearization, further reduce computation.
Experiments on 2D and 3D tasks with up to 1024 agents
demonstrate improved exploration and constraint satisfaction over state-of-the-art derivative-based distributed optimization, and superior scalability to existing distributed sampling methods.
\end{abstract}


\section{Introduction}

Cooperative trajectory planning is fundamental to the safe and efficient operation of robot teams in shared environments, such as aerial swarms navigating clutter~\cite{zhou2021egoswarm}, autonomous vehicles traversing intersections~\cite{ferranti2023distributed}, and warehouse fleets transporting goods~\cite{wurman2008coordinating}. 
In such settings, robots must plan trajectories that optimize task performance subject to their dynamics, actuation limits, and collision avoidance with obstacles and other robots.
These requirements render planning non-convex, coupling decisions through nonlinear dynamics across the horizon and inducing competing local minima among routes and passing orders.
At the same time, scale compounds the difficulty: the computational cost of coordinating these coupled trajectories grows rapidly with team size.
Meeting these requirements calls for methods that jointly address scalability and non-convexity.

Coordination has been pursued through multi-agent path finding \cite{sharon2012conflict}, reactive feedback methods \cite{wang2017safety}, learned policies \cite{chen2017decentralized}, and dynamic games \cite{lecleach2020algames}; however, these approaches do not jointly optimize the team's trajectories.
In contrast, trajectory optimization (TO) plans all agents' trajectories through a joint optimal control problem, yet solving that centrally quickly becomes prohibitive as teams grow.
Distributed approaches address this bottleneck by decomposing the full problem into local agent-level subproblems solved in parallel and coordinated through neighborhood communication.
Some of the most prominent ones build on the alternating direction method of multipliers (ADMM) \cite{boyd2011distributed}, a well-established distributed optimization framework.
Such methods have enabled 
distributed model predictive control \cite{summers2012distributed, saravanos2024dmpcs}, collaborative manipulation \cite{shorinwa2020scalable} and legged locomotion \cite{zhou2026aclm}, 
as well as distributed TO architectures scaling to thousands of agents \cite{saravanos2023distributed, saravanos2026distributedcs}.
Nevertheless, these methods typically rely on derivative-based local solvers, remaining prone to initialization and poor local minima.

%
    
    \begin{figure}[!t]
        \centering
        \includegraphics[width=\linewidth]{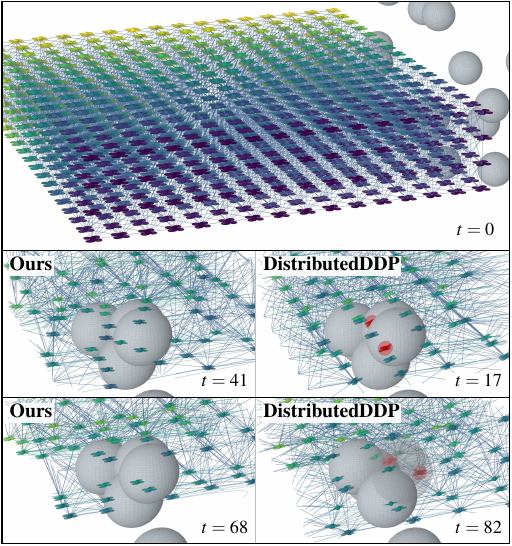}
        \vspace{-7mm}
        \caption{\textbf{Scalable multi-agent coordination with DS-TO~(Ours).} By keeping the sampled control dimension independent of neighborhood size, DS-TO enables 1024 locally coupled agents to navigate strongly non-convex interactions while maintaining collision-free coordination beyond the local minima encountered by existing gradient-based distributed methods.}
        \vspace{-8mm}
        \label{fig:traj_drone}
    \end{figure}
    

Sampling-based derivative-free methods such as model predictive path integral (MPPI) \cite{williams2017information} and the cross entropy method (CEM) \cite{kobilarov2012cross} are natural candidates for such non-convex landscapes.
Although such methods are widely used in the single-agent setting \cite{gandhi2021robust, yin2023shield}, 
their sample efficiency deteriorates as the search space grows, fundamentally limiting their extension to large-scale multi-agent problems. 
Some notable attempts include ADMM schemes that sample joint neighborhood trajectories \cite{wang2022sampling, yoon2025distributed}, belief propagation for distributed inference \cite{jiang2024distributed}, and decentralized MPPI accounting for neighbors through predicted trajectories \cite{dergachev2025corl}.


The effectiveness of distributed sampling schemes hinges on two factors: the dimension of each local search and the safety information available to evaluate rollouts.
On one side, jointly sampling ego and neighboring controls \cite{wang2022sampling,yoon2025distributed} exposes the sampler to all constraints but inflates the search dimension.
On the other hand, splittings that isolate agent dynamics \cite{saravanos2023distributed,chen2023multiagent} would keep the sampled dimension small but hide the constraints inside auxiliary copies, leaving the sampler to explore a surrogate objective blind to the true safety landscape.
Some decentralized MPPI schemes \cite{dergachev2025decentralized}
avoid the trade-off entirely, but only by abandoning coordination through a joint problem, treating neighbors as fixed predictions rather than co-optimizing agents.
The core challenge is therefore to retain direct exploration of the non-convex safety landscape and joint trajectory coordination while sampling only each agent's own controls.

%
%


We propose \textbf{\underline{D}istributed \underline{S}ampling-based \underline{T}rajectory \underline{O}ptimization (DS-TO)}, an ADMM-based distributed sampling-based TO framework that retains constraint-aware exploration while sampling only each agent’s own controls. 
The proposed formulation separates dynamically feasible primal trajectories from local copies subject to safety constraints, with global consensus variables enforcing consistency between the two. 
%
%
Soft constraint penalties in the primal objective let stochastic optimizers explore the non-convex safety landscape directly, while the local copies coordinate constraint satisfaction without entering the sampled search, keeping its dimension independent of neighborhood size.
%
Instantiating the primal solver with MPPI, our experiments demonstrate strong constraint satisfaction, improved exploration in non-convex scenarios, and favorable scaling relative to state-of-the-art distributed TO baselines (Fig. \ref{fig:traj_drone}).

    The contributions of this work are:
    \begin{itemize}
        \item We formulate a distributed TO decomposition that separates agent dynamics from local constraints through consensus, enabling parallel agent updates and feasibility subproblems that decompose across time.
        \item We introduce DS-TO, a scalable, fully decentralized algorithm tailored to stochastic solvers, achieving constraint-aware exploration while keeping the sampled control dimension independent of neighborhood size.
        \item Extensive evaluation on 2D and 3D tasks with up to 1024 agents, demonstrating improved exploration and constraint satisfaction over state-of-the-art derivative-based distributed optimization, and superior scalability than existing distributed sampling methods.
    \end{itemize}

\section{Preliminaries}

    \subsection{Notation}
        Scalars and vectors use lowercase letters, \textit{e.g.}, $a\in\mathbb{R}$, $v\in\mathbb{R}^n$, respectively; matrices use uppercase, \textit{e.g.}, $A\in\mathbb{R}^{m\times n}$; and calligraphic letters denote sets, with $|\mathcal{S}|$ their cardinality. We use $(\cdot)^\top$ for transpose, $I$ for identity, $\operatorname{blkdiag}(\cdot)$ for block-diagonal concatenation, and $\|\cdot\|$ for the Euclidean norm, with $\|v\|_{A}^2\triangleq v^\top A v$ as the weighted norm. A tilde $(\tilde{\cdot})$ denotes a local copy and a bar $(\bar{\cdot})$ a global variable. $\mathcal{U}(a,b)$ denotes the uniform distribution on $[a,b]$, and $\mathcal{N}(\mu,\Sigma)$ a Gaussian distribution with mean $\mu$ and covariance $\Sigma$.
        
    \subsection{Problem Definition}

        We consider a team of $N$ agents, with agent set $\mathcal{V}\triangleq\{1,\ldots,N\}$, over a finite horizon of $T$ steps. At time $t$, agent $i\!\in\!\mathcal{V}$ has state $x_{i,t}\in\mathbb{R}^{n_x}$ and control input $u_{i,t}\in\mathbb{R}^{n_u}$. Each agent $i$ has a neighborhood set $\mathcal{N}_i\!\subseteq\!\mathcal{V}\setminus\!\{i\}$, comprising agents with which it exchanges trajectory information.
        
        We formulate the problem as an optimal control problem (OCP) over the agents' trajectories. For agent $i$, let ${x}_i\triangleq({x}_{i,0},\ldots,{x}_{i,T})$ and ${u}_i\triangleq({u}_{i,0},\ldots,{u}_{i,T-1})$, with ${X}\triangleq({x}_1,\ldots,{x}_N)$ and ${U}\triangleq({u}_1,\ldots,{u}_N)$ denoting the team trajectories. The stage cost $J_{i,t}$ captures task performance and control effort, while $J_{i,T}$ encodes the terminal goal. The dynamics $f_{i,t}$ govern state evolution, $g_{i,t}$ captures agent-wise requirements (obstacle avoidance and state/control limits), and $h_{ij,t}$ pairwise safety requirements (minimum separation). For all $i\in\mathcal{V}$ and $t=0,\dots,T-1$, the resulting OCP is
            \begin{subequations}\label{eq:ocp}
            \begin{align}
            \min_{X, U}\quad
            & \sum_{i\in\mathcal{V}}
              \left[
              \sum_{t=0}^{T-1} J_{i,t}(x_{i,t},u_{i,t})
              + J_{i,T}(x_{i,T})
              \right]
              \label{eq:ocp_obj} \\
            \mathrm{s.t.}\quad
            & x_{i,0}=x_{i,\mathrm{init}}, \quad x_{i,t+1}
              =f_{i,t}(x_{i,t},u_{i,t}),
              \label{eq:ocp_dyn} \\
            & g_{i,t}(x_{i,t},u_{i,t})\leq 0,
              \label{eq:ocp_local} 
              \\
            & h_{ij,t}(x_{i,t},x_{j,t}, u_{i,t},u_{j,t})\leq 0,
              \quad j\in\mathcal{N}_i
              \label{eq:ocp_pairwise}
            \end{align}
            \end{subequations}
         where the terminal constraints can be handled analogously.
            
    \subsection{Alternating Direction Method of Multipliers (ADMM)}

        We briefly review the standard two-block ADMM \cite{boyd2011distributed}. Consider the constrained optimization problem
            \begin{equation}
            \label{eq:admm_problem}
            \min_{p, q} \quad
            \phi(p)+\psi(q)
            \quad \mathrm{s.t.} \quad
            Ap + Bq= c,
            \end{equation}
        where $p\in\mathbb{R}^{n_p}$ and $q\in\mathbb{R}^{n_q}$ are the two primal blocks, $\phi$ and $\psi$ are their objective terms, and $A$, $B$, and $c$ define the coupling constraint. Introducing the dual variable $\lambda$ and penalty parameter $\rho>0$, the augmented Lagrangian (AL) is
            \begin{equation}
            \label{eq:admm_al}
            \mathcal{L}_{\rho}(p,q,\lambda)
            \!=\!
            \phi(p)\!+\!\psi(q)\!
            +\!\lambda^{\top}\!r\!
            +\!\frac{\rho}{2}\|r\|^2,
            \quad\!\!\!
            r\!\triangleq\!
            Ap + Bq - c
            \end{equation}
        At iteration $\ell$, ADMM alternates between minimization over the two primal blocks followed by a dual ascent step,
            \begin{subequations}\label{eq:admm_updates}
            \begin{align}
            p^{\ell+1}
            &=\arg\min_{p}\,
            \mathcal{L}_{\rho}
            (p, q^{\ell},\lambda^{\ell}),
            \label{eq:admm_p}\\
            q^{\ell+1}
            &=\arg\min_{q}\,
            \mathcal{L}_{\rho}
            (p^{\ell+1},q,\lambda^{\ell}),
            \label{eq:admm_q}\\
            \lambda^{\ell+1}
            &=\lambda^{\ell}
            +\rho\!\left(
            Ap^{\ell+1}
            +Bq^{\ell+1}
            -c\right).
            \label{eq:admm_dual}
            \end{align}
            \end{subequations}
        The algorithm holds similarly for multiple primal blocks.
    
    \subsection{Model Predictive Path Integral Control}

        Model Predictive Path Integral (MPPI) control is a derivative-free method for optimal control that iteratively updates a nominal control sequence using importance-weighted, dynamically feasible rollouts generated by forward simulation \cite{williams2017information}. Given a nominal sequence $u$, MPPI samples $K$ perturbed control sequences around it according to
            \begin{equation}
            u^{(k)}_t=u_t+\epsilon^{(k)}_t,
            \qquad
            \epsilon^{(k)}_t\sim\mathcal{N}(0,\Sigma),
            \end{equation}
        where $\Sigma$ is the sampling covariance. Each perturbed sequence is propagated through the dynamics and assigned a trajectory cost $J^{(k)}$. The samples are then weighted as
            \begin{equation}
            w_k=\exp(-\alpha J^{(k)}) / \bar{w}, \qquad \bar{w} = \sum_{r=1}^{K}\exp(-\alpha J^{(r)}),
            \end{equation}
        where $\alpha>0$ controls the concentration toward lower-cost rollouts. The nominal control trajectory, corresponding to the mean of the sampling distribution, is updated according to
            \begin{equation}
            u_t\leftarrow
            u_t+\sum_{k=1}^{K}w_k\epsilon^{(k)}_t,
            \quad t=0,\ldots,T-1.
            \end{equation}
            %

\section{Methodology}

    DS-TO is built around a sampling-tailored decomposition: each agent samples only its own controls, while dynamics-free local copies represent neighboring trajectories and explicitly enforce local and interagent constraints. Soft penalties expose these constraints directly to the sampled primal rollouts, while ADMM consensus reconciles the dynamically feasible primal trajectories with their local copies. This keeps the sampled dimension independent of neighborhood size while allowing direct exploration of the non-convexity.

    \subsection{Consensus Reformulation and Augmented Lagrangian}

        Let $\mathcal{N}_i^+=\mathcal{N}_i\cup\{i\}$ denote agent $i$'s augmented neighborhood. Agent $i$ maintains local state and input copies $\tilde{x}_{j,t}^{i},\tilde{u}_{j,t}^{i}$ for each $j\in\mathcal{N}_i^+$, where superscript $i$ identifies the agent holding the copy. We stack these variables as $\tilde{x}_{i,t}^{a}:=(\tilde{x}_{j,t}^{i})_{j\in\mathcal{N}_i^+}$ and $\tilde{u}_{i,t}^{a}:=(\tilde{u}_{j,t}^{i})_{j\in\mathcal{N}_i^+}$, with superscript $a$ denoting quantities stacked over $\mathcal{N}_i^+$. The local copies provide the variables on which constraints are imposed: $g_{i,t}$ acts on agent $i$'s own copy, while $h_{i,t}^{a}$ collects the interagent constraints with $j\in\mathcal{N}_i$. Global variables $\bar{x}_{i,t},\bar{u}_{i,t}$ provide a common representation of each agent's state and input, with $\bar{x}_{i,t}^{a}$ and $\bar{u}_{i,t}^{a}$ stacking the corresponding global variables over $\mathcal{N}_i^+$. Defining $Z=(X,U)$, $\tilde{Z}^{a}=(\tilde{X}^{a},\tilde{U}^{a})$, and $\bar{Z}=(\bar{X},\bar{U})$ we reformulate the OCP \eqref{eq:ocp} as follows
        \begin{equation}
        \label{eq:consensus_reform}
        \begin{aligned}
            \min_{Z,\tilde{Z}^a,\bar{Z}}\quad
            & \sum_{i\in\mathcal{V}}\Phi_i(Z_i,\tilde{Z}_i^a) \\
            \text{s.t.}\quad
            & x_{i,0}=x_{i,\mathrm{init}}, \quad
              x_{i,t+1}=f_{i,t}(x_{i,t},u_{i,t}), \\
            & g_{i,t}(\tilde{x}_{i,t}^{i},\tilde{u}_{i,t}^{i})\leq 0, \quad
              h_{i,t}^{a}(\tilde{x}_{i,t}^{a},\tilde{u}_{i,t}^{a})\leq 0, \\
            & x_{i,t}=\bar{x}_{i,t}, \quad
              u_{i,t}=\bar{u}_{i,t}, \\
            & \tilde{x}_{i,t}^{a}=\bar{x}_{i,t}^{a}, \quad
              \tilde{u}_{i,t}^{a}=\bar{u}_{i,t}^{a},
        \end{aligned}
        \end{equation}
        where the agent-wise objective $\Phi_i$ combines task cost with soft penalties that expose constraint-induced non-convexity and neighboring-trajectory information to the primal update: 
        %
        \begin{align}
        \Phi_i(Z_i,\tilde{Z}_i^a)
        \triangleq\;&
        \sum_{t=0}^{T-1}J_{i,t}(x_{i,t},u_{i,t})
        +J_{i,T}(x_{i,T})
        \label{eq:local_objective}\\
        &+\sum_{t=0}^{T-1}
        \beta_g\,c\!\left(g_{i,t}(x_{i,t},u_{i,t})\right)
        \nonumber\\
        &+\sum_{t=0}^{T-1}\sum_{j\in\mathcal{N}_i}
        \beta_h\,c\!\left(
        h_{ij,t}(x_{i,t},u_{i,t},
        \tilde{x}_{j,t}^{i},\tilde{u}_{j,t}^{i})
        \right),
        \nonumber
        \end{align}
        where the penalty $c(\cdot)$ is zero when constraints are satisfied and positive otherwise, with $\beta_g,\beta_h>0$. At consensus, feasibility of the local copies ensures the original constraints in~\eqref{eq:ocp} are satisfied and therefore the soft penalties vanish.

        To form the Augmented Lagrangian~(AL), we introduce dual variables $\lambda_i$ and $\tilde{\lambda}_i^a$ for $Z_i\!=\!\bar{Z}_i$ and $\tilde{Z}_i^a\!=\!\bar{Z}_i^a$, respectively. Define $R:=\operatorname{blkdiag}(\rho I,\mu I)$ and $\tilde{R}_i:=|\mathcal{P}_i|^{-1}\operatorname{blkdiag}(\tilde{\rho}I,\tilde{\mu}I)$, where $\mathcal{P}_i:=\{k\mid i\in\mathcal{N}_k^+\}$ collects the agents holding copies of agent $i$, and all penalty coefficients are positive. Normalizing by $|\mathcal{P}_i|$ preserves the primal's contribution as the copy count increases. The augmented penalty is $\tilde{R}_i^a:=\operatorname{blkdiag}((\tilde{R}_j)_{j\in\mathcal{N}_i^+})$. The AL is
        \begin{equation}
        \begin{aligned}
        \label{eq:AL}
            \mathcal{L}
            = \sum_{i \in \mathcal{V}} \mathcal{L}_i =& \sum_{i \in \mathcal{V}} \Bigg[\Phi_i(Z_i,\tilde{Z}_i^a) + I_f(Z_i) + I_c(\tilde{Z}_i^a) \\
            & +\! \lambda_i^\top\!(Z_i-\bar{Z}_i)
            \!+\! \frac{1}{2}\|Z_i-\bar{Z}_i\|_{R}^{2} \\
            & +\! (\tilde{\lambda}_i^a)^\top\!
            (\tilde{Z}_i^a-\bar{Z}_i^a)
            \!+\! \frac{1}{2}
            \|\tilde{Z}_i^a-\bar{Z}_i^a\|_{\tilde{R}_i^a}^{2}
            \Bigg]
        \end{aligned}
        \end{equation}
        where $I_f(Z_i)$, $I_c(\tilde{Z}_i^a)$ are indicators of dynamically feasible primal and constraint-feasible local-copy sets in~\eqref{eq:consensus_reform}.
    
        In the following, we apply ADMM to alternate between local updates over $(Z_i,\tilde{Z}_i^a)$, decentralized consensus updates over $\bar{Z}_i$, and dual updates over $(\lambda_i,\tilde{\lambda}_i^a)$, corresponding to the updates in~\eqref{eq:admm_updates}.
        Since the augmented Lagrangian is separable across agents, the local updates can be performed in parallel.

    \subsection{\blockagentwise: Agent-wise Local Subproblem}
    \label{subsec:blockagentwise}
        In the first ADMM block, each agent independently minimizes $\mathcal{L}_i$ over its primal and local-copy variables $(Z_i,\tilde{Z}_i^a)$, while keeping the global and dual variables fixed.
        The optimization variables remain coupled through the soft penalty terms $\beta_h c(h_{ij,t})$ in $\Phi_i$. Jointly solving for these variables cannot exploit distinct structures: the primal variables are constrained only by the dynamics, whereas the local copies are subject to hard constraints but timewise independent. We therefore use a block-coordinate update to separate updates, yielding subproblems tailored to their respective structures.

        The block-coordinate update can follow either a Jacobi or Gauss--Seidel~(GS) scheme; we use Jacobi by default, allowing both blocks to use previous iterates and perform the computationally intensive updates in parallel within each agent. The ADMM penalties maintain coupling between the two blocks through the shared consensus variables.
        Suppressing the inner block-coordinate iteration index, underlined variables denote the latest available values. They are initialized as $\{\underline{x}_{i,t},\underline{u}_{i,t}\}=\{x_{i,t}^{\ell},u_{i,t}^{\ell}\}$ and $\underline{\tilde{Z}}_i^{a}=\tilde{Z}_i^{a,\ell}$.


    \subsubsection{\blocktraj: Primal Trajectory Update}
    \label{subsubsec:blocktraj}
        Fixing the local copy $\underline{\tilde{Z}}_i^a$, each agent $i$ optimizes only for $Z_i$:
        \begin{align}
        Z_i^{\ell+1}
        = & \argmin_{Z_i}\;
        \Phi_i\!\left(Z_i,\underline{\tilde{Z}}_i^{a}\right)
        +\frac{1}{2}
        \left\|
        Z_i-\bar{Z}_i^\ell+R^{-1}\lambda_i^\ell
        \right\|_R^2
        \notag\\
        & \mathrm{s.t.}\;
        x_{i,0}=x_{i,\mathrm{init}}, \quad x_{i,t+1}=f_{i,t}(x_{i,t},u_{i,t}),
        \label{eq:subblock_A}
        \end{align}
        where $I_f(Z_i)$ is enforced by the initial condition and dynamics, reducing the problem via forward rollout to unconstrained optimization over the control sequence.
        This is well suited to sampling-based optimization, since samples are drawn only over agent $i$'s controls, with a fixed sampling dimension of $T\!\times\!n_u$ independent of neighborhood size. The soft penalties in $\Phi_i$ incorporate constraint and neighboring-trajectory information, enabling exploration of the resulting non-convex landscape. Since the constraint-feasible local copies need not satisfy the dynamics, these penalties guide the primal trajectory toward consistency with them.
        
    \subsubsection{\blocksafe: Local-Copy Feasibility Update}
    
    Since $\tilde{Z}_i^a$ carries no dynamic constraints, its update decomposes across time. With $\{\underline{x}_{i,t},
        \underline{u}_{i,t}\}$ fixed, agent $i$ solves
    \begin{align}
        \tilde{Z}_{i,t}^{a,\ell+1}
        = \argmin_{\tilde{Z}_{i,t}^{a}}\;&
        \frac{1}{2}
        \left\|
        \tilde{Z}_{i,t}^{a}
        -\bar{Z}_{i,t}^{a,\ell}
        +(\tilde{R}^a_{i,t})^{-1}\tilde{\lambda}_{i,t}^{a,\ell}
        \right\|_{\tilde{R}^a_{i,t}}^2
        \notag\\
        &+
        \sum_{j\in\mathcal{N}_i}
        \beta_h c\!\left(
        h_{ij,t}(
        \underline{x}_{i,t},
        \underline{u}_{i,t},
        \tilde{x}_{j,t}^{i},
        \tilde{u}_{j,t}^{i})
        \right)
        \notag\\
        \text{s.t.}\quad
        g_{i,t}(\tilde{x}_{i,t}^{i},&\tilde{u}_{i,t}^{i})\leq 0,\quad h_{i,t}^{a}(\tilde{x}_{i,t}^{a},\tilde{u}_{i,t}^{a})\leq 0,
        \label{eq:subblock_B}
    \end{align}
    for each $t=0,\ldots,T-1$.
    Unlike the primal update, this subproblem explicitly enforces constraints $g_i$ and $h_{i}^a$, thereby producing constraint-feasible local copies. The decomposition confines neighborhood-dependent scaling to each time step rather than over the entire horizon.
    
    If $\beta_h=0$, the primal and local-copy updates decouple within the ADMM primal step~\eqref{eq:admm_p}. The soft penalties introduce this coupling, allowing constraint-feasible local copies to inform the dynamically feasible primal update and thereby motivating the block-coordinate update scheme.
    
    


    \subsection{\blockglobal: Global Consensus and Dual Updates}
    \label{subsec:blockglobal}
    Minimizing $\mathcal{L}_i$ over $\bar{Z}_i$ yields a second ADMM block with penalty weights $R,\tilde{R}_i$ and dual corrections $\lambda_i,\{\tilde{\lambda}_i^k\}_{k\in\mathcal{P}_i}$,
        \begin{equation}
        \label{eq:global_update}
            \bar{Z}_i^{\ell+1}
            \!\!=\!
            \big(R\!+\!|\mathcal{P}_i|\tilde{R}_i\big)^{-1}
            \Bigg[
            R Z_i^{\ell+1} \!\!\!+\! \lambda_i^\ell 
            \!+\!\!\! \sum_{k\in\mathcal{P}_i}\!\!
            \left(
            \tilde{R}_i\tilde{Z}_i^{k,\ell+1}
            \!\!+\!\tilde{\lambda}_i^{k,\ell}
            \right)
            \!\!\Bigg]
        \end{equation}
    Agent $i$ evaluates~\eqref{eq:global_update} using $(Z_i,\lambda_i)$ and $(\tilde{Z}_i^k,\tilde{\lambda}_i^k)$ received from agents $k\in\mathcal{P}_i$. Since $|\mathcal{P}_i|\tilde{R}_i=\operatorname{blkdiag}(\tilde{\rho}I,\tilde{\mu}I)$, the total weight assigned to all local copies of agent $i$, $\{\tilde{Z}_i^k\}_{k\in\mathcal{P}_i}$, remains fixed as $|\mathcal{P}_i|$ grows, preserving the relative influence of $Z_i$ in the consensus update.
    
    Finally, each agent updates its dual variables $\lambda_i$ and $\tilde{\lambda}_i^a$ by ascent on the primal and local copy consensus residuals,
    \begin{align}
        \lambda_i^{\ell+1}
        &=\lambda_i^\ell
        +R(Z_i^{\ell+1}-\bar{Z}_i^{\ell+1}),\\
        \tilde{\lambda}_i^{a,\ell+1}
        &=\tilde{\lambda}_i^{a,\ell}
        +\tilde{R}_i^a(\tilde{Z}_i^{a,\ell+1}-\bar{Z}_i^{a,\ell+1}),
    \end{align}
    where $\bar{Z}_i^{a,\ell+1}$ collects $\bar{Z}_j^{\ell+1}$ for $j\in\mathcal{N}_i^+$, with each neighbor $j\in\mathcal{N}_i$ communicating the updated $\bar{Z}_j^{\ell+1}$ to agent $i$.
\section{Implementation Decisions}

    We describe the solver choices and approximations used in DS-TO. These choices are specific to our implementation, and alternatives can be used within the same framework.
    
    \subsection{MPPI for Constraint-Aware Exploration}

        We employ MPPI to solve~\eqref{eq:subblock_A}, sampling each agent's control sequence~$T \times n_u$ and generating dynamically feasible rollouts in parallel. The soft penalties $\beta_g c(g_{i,t})$ and $\beta_h c(h_{ij,t})$ guide dynamically feasible exploration of the non-convex constraint landscape. Penalty evaluations can be performed in parallel across samples and constraints without requiring any constraint derivatives. Note that a convergence guarantee accounting for the inexact stochastic MPPI updates lies beyond the scope of this work.
    


    \subsection{Constraint Linearization and Relaxation}

        The local copy variables are warm-started from the global consensus variables, which also serve as the proximal targets in~\eqref{eq:subblock_B} and carry information from the preceding primal update through~\eqref{eq:global_update}. This motivates normalizing the local-copy penalties to preserve the influence of explored primal trajectories.
        Although~\eqref{eq:subblock_B} decomposes across time, fully solving each nonlinear subproblem requires multiple iterations. We instead linearize the constraints and omit the interagent soft penalty $\beta_h c(h_{ij,t})$, yielding a small QP independent of the current primal update and relinearize at the next ADMM iteration. Since the omitted penalty and its gradient vanish at consensus, this does not alter the stationarity conditions.

        Since global consensus may violate the constraints, linearization around consensus variables can yield an infeasible subproblem. We therefore introduce slack variables initialized from the current violations to ensure a feasible relaxed subproblem. We do not warm-start the local solver dual variables, avoiding bias toward the previous active set.

            \begin{figure*}[t]
                \centering
                \includegraphics[width=\textwidth]{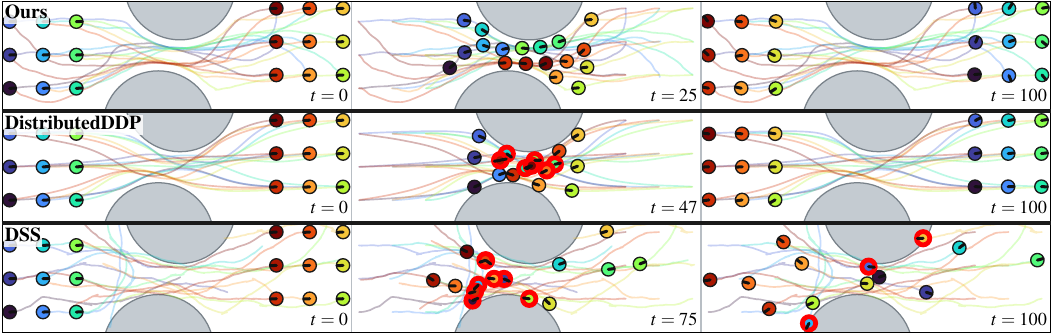}
                \vspace{-7mm}
                \caption{Trajectories in the Gate with $18$ agents. The middle snapshot is taken when agents are closest to one another. Red agent edges indicate collisions.
                }
            \label{fig:traj_gate}
            \vspace{-6mm}
            \end{figure*}

\section{Experiments}
\label{sec:experiments}

    We first study \textit{fully connected} ($|\mathcal{N}_i^+| =N, \forall i\in\mathcal{V}$) agents in tasks emphasizing interagent coordination and non-convex route exploration, then evaluate \textit{locally connected} ($|\mathcal{N}_i^+| \ll N, \forall i\in\mathcal{V}$) agents to assess scalability. The agents are modeled as unicycles or drones.
    We select baselines with complementary characteristics. Merged \textit{DistributedDDP}~\cite{saravanos2023distributed} is a scalable deterministic baseline using three-block ADMM with primal--local and local--global consensus and gradient-based trajectory updates. DSS~\cite{wang2022sampling,yoon2025distributed} utilizes MPPI with soft penalties using two-block ADMM, but samples control trajectories over the augmented neighborhood, increasing the search dimension to~$|\mathcal{N}_i^+| \times T \times n_u$. We also include SQP~\cite{nocedal2006numerical}, with OSQP~\cite{osqp} as the internal solver. The ablations of DS-TO either remove the interagent soft penalty ($\beta_h=0$), both of them ($\beta_g,\beta_h=0$), or replace the default parallel Jacobi update with sequential Gauss--Seidel.

    Distributed methods use 100 ADMM iterations, while SQP uses 100 outer iterations. For DS-TO, $\rho,\mu,\tilde{\rho},\tilde{\mu}=10$, with local-copy penalties scaled by $1/|\mathcal{P}_i|$. Both DS-TO and DSS use $c(\cdot)=[\max(0,\cdot)]^2$ with $\beta_g=\beta_h=10^5$. DS-TO performs one Jacobi block-coordinate update per ADMM iteration. For unicycle tasks, baseline penalties are tuned over $\rho=\mu\in\{0.1, 1, 10, 100, 1000\}$; DSS additionally sweeps $\alpha \in \{10^2, 10^3, 10^4\}$. For each task and scale, We report the best \textit{Clean} or \textit{Succ} rate, as defined in the corresponding sections. For drone experiments, we use the same ADMM penalty as DS-TO.

    DistributedDDP performs eight DDP updates per ADMM iteration, while DS-TO and DSS perform three MPPI updates with $K=4096$ samples. DS-TO uses $\alpha=100$, while DSS uses the tuned value above. The sampling covariances are $\Sigma_{\mathrm{uni}}\!=\!\operatorname{diag}(0.4,0.2)$ and $\Sigma_{\mathrm{drone}}\!=\!\operatorname{diag}(10^{-2}\!,10^{-4}\!,10^{-4}\!,10^{-4})$. For sampling methods, each environment and scale is evaluated over five random seeds.

    The agents' dynamics are discretized by forward Euler using $\Delta t=0.1\,\mathrm{s}$. The unicycle has state $x_i=[p_{x,i},p_{y,i},\theta_i]^\top\in\mathbb{R}^3$ (planar position and heading) and control $u_i=[\upsilon_i,\dot{\theta}_i]^\top\in\mathbb{R}^2$ (linear and angular velocity). The drone follows the model in~\cite{luukkonen2011modelling}, with state $x_i=[p_i^\top,\dot p_i^\top,\eta_i^\top,\Omega_i^\top]^\top\in\mathbb{R}^{12}$ (position, velocity, Euler angles, and body angular velocity) and control $u_i=[\mathcal{F}_i,\tau_i^\top]^\top\in\mathbb{R}^4$ (total thrust and body torque). We use mass $m=1\,\mathrm{kg}$ and inertia tensor $\mathcal{I}=\operatorname{diag}(0.02,0.02,0.04)\,\mathrm{kg\!\cdot\!m^2}$. Control bounds $\upsilon_i\in[-1,3]$, $|\dot{\theta}_i|\leq2$, $\mathcal{F}_i\in[0,20]$, $|\tau_{i,k}|\leq0.2$ are enforced by clamping in MPPI and box constraining the QPs. Since DistributedDDP does not enforce these bounds, control bound violations are excluded from comparison metrics.

    All tasks use per-agent quadratic stage
    $J_{i,t}(x_{i,t},u_{i,t}) = \tfrac12\|x_{i,t}-x_i^{\mathrm{ref}}\|_Q^2+\tfrac12\|u_{i,t}-u^{\mathrm{ref}}\|_R^2$
    and terminal cost
    $J_{i,T}(x_{i,T}) = \tfrac12\|x_{i,T}-x_i^{\mathrm{ref}}\|_{Q_T}^2$,
    where $x_i^{\mathrm{ref}}$ is the target state. For unicycle, $Q=\operatorname{diag}(1,1,0)$, $R=\operatorname{diag}(3,0.3)$, and  $u^{\mathrm{ref}}=0$; for drones, $Q=\operatorname{blkdiag}(I_3,0.1I_3,0_{3\times3},0.01I_3)$, $R=0.01I_4$, and $u^{\mathrm{ref}}=[9.81,0,0,0]^\top$, where both use $Q_T=2400 \times Q$.
    State trajectories are initialized at $x_{i,\mathrm{init}}$ and control sequences at $u^{\mathrm{ref}}$.
    Obstacle constraints are defined as $g_{i,t}^{\mathrm{obs}}=(r_{\mathrm{agent}}+r_{\mathrm{obs}}) - \|p_{i,t}-c_{\mathrm{obs}}\|$, where $c_{\mathrm{obs}}$ and $r_{\mathrm{obs}}$ denote the obstacle center and radius, respectively; the resulting constraints are stacked into $g_{i,t}$.
    Interagent collision avoidance is imposed as $h_{ij,t}=2r_{\mathrm{agent}} - \|p_{i,t}-p_{j,t}\|$.


    Experiments were conducted on an Intel i9-13900K CPU, 64\,GB RAM, and an NVIDIA RTX 4090 GPU. ADMM methods execute on the GPU while SQP (OSQP) on the CPU. For Jacobi, parallel runtime is estimated as the maximum solve time of~\eqref{eq:subblock_A} and~\eqref{eq:subblock_B}, measured sequentially with GPU synchronization. 
    For locally connected cases with $N\geq64$, methods are serialized in batches of 32 agents due to hardware limits; we report average per-batch runtime to reflect parallel execution in a distributed deployment.

    \vspace{-1mm}

    \subsection{Fully Connected Agents}
    \label{subsec:FCA}
        We employ unicycles of $r_{\mathrm{agent}}=0.3\,\mathrm{m}$ and evaluate 20 environments. \textit{Clean} denotes the fraction of episodes in which all agents finish within $2r_{\mathrm{agent}}$ of their goals while all obstacle and interagent violations remain below $0.05\,\mathrm{m}$. We also report the maximum penetration $\delta_{\max}$ over all episodes.

        \begin{figure}[b]
        \vspace{-4mm}  
        \includegraphics[width=0.48\textwidth]{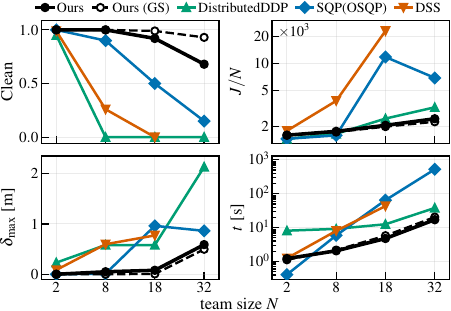}
        \vspace{-2mm}
        \caption{Comparison in the Gate. Top left: \textit{Clean}~($\uparrow$); bottom left: $\delta_{\max}$~($\downarrow$); top right: $J/N$~($\downarrow$); bottom right: computation time~($\downarrow$).}
        \label{fig:fig_gate}
        \vspace{-0.5mm}
        \end{figure}

            \begin{figure*}[t]
                \centering
                \includegraphics[width=\textwidth]{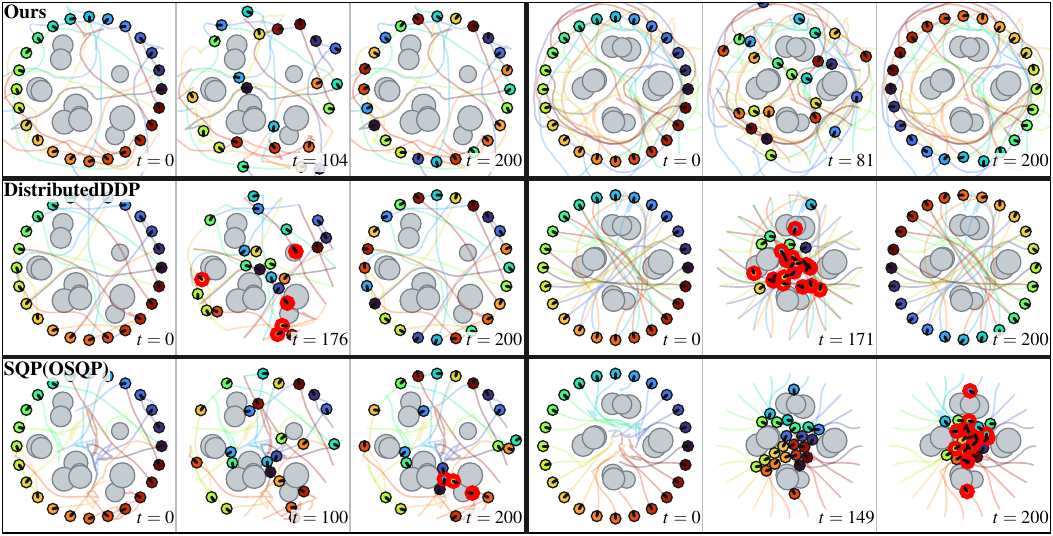}
                \vspace{-7mm}
                \caption{Trajectories in the Circle with $24$ agents. The left two columns show the \textit{antipodal} goal assignment, and the right two columns show the random \textit{permutation} assignment. The middle snapshot is taken when agents are closest to one another. Red agent edges indicate collisions.}
            \label{fig:traj_circle}
            \vspace{-6mm}
            \end{figure*}

        \subsubsection{Gate}

            This task requires two groups to exchange positions through a narrow passage. We consider $N\in\{2,8,18,32\}$ with $T=100$, pairing corresponding positions to equalize travel distances and offsetting initial heights by $\pm0.3\,\mathrm{m}$ to break symmetry. The two obstacles share $r_{\mathrm{obs}}\sim\mathcal{U}(0.5,3.0)\,\mathrm{m}$, with horizontal offsets from $\mathcal{U}(-1,1)\,\mathrm{m}$; passage width and vertical center are sampled from $\mathcal{U}(1.3,2.4)\,\mathrm{m}$ and $\mathcal{U}(-0.6,0.6)\,\mathrm{m}$. Large obstacles force coordination through the central passage, as shown in Fig.~\ref{fig:traj_gate}, whereas smaller ones allow detours.

            Figure~\ref{fig:fig_gate} shows that DS-TO achieves \textit{Clean} rates of $1.00$, $0.92$, and $0.68$ at $N=8,18,32$, respectively. In contrast, DistributedDDP has no clean episodes for $N\geq8$, while SQP drops to $0.15$ at $N=32$, highlighting the limited exploratory capability of gradient-based updates. DSS also degrades with team size and exceeds memory at $N=32$. DS-TO scales more favorably, requiring $17.1\,\mathrm{s}$ at $N=32$, versus $38.0\,\mathrm{s}$ for DistributedDDP and $523.3\,\mathrm{s}$ for SQP, while achieving a lower $J/N$ of $2.44\times10^3$, compared with $3.27\times10^3$ and $6.92\times10^3$, respectively. We omit $J/N$ hereafter. 

            Finally, we compare the default Jacobi update with GS. 
            GS improves \textit{Clean} from $0.92$ to $0.99$ at $N=18$ and from $0.68$ to $0.93$ at $N=32$, at up to $21\%$ additional computation time. The modest subproblem sizes, however, limit the computational benefit of parallel Jacobi updates.

        \subsubsection{Circle}

            \begin{figure}[h]
            \vspace{-1.5mm}
            \includegraphics[width=0.48\textwidth]{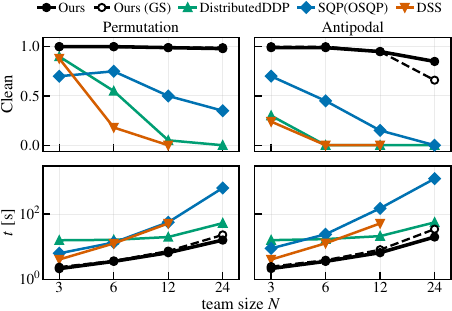}
            \vspace{-2mm}
            \caption{Comparison in the Circle. The left and right columns show the \textit{permutation} and \textit{antipodal} assignments, respectively, while the top and bottom rows report the \textit{Clean} rate~($\uparrow$) and log-scale computation time~($\downarrow$).}
            \label{fig:fig_circle}
            \vspace{-3mm}
            \end{figure}
            
            \begin{figure}[b!]
            \vspace{-4mm}
            \includegraphics[width=0.48\textwidth]{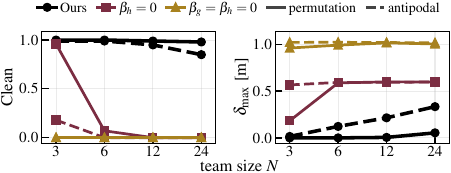}
            \vspace{-2mm}
            \caption{Soft penalty ablations in the Circle. Left: \textit{Clean} rate~($\uparrow$); right: maximum penetration~$\delta_{\max}$~($\downarrow$).}
            \label{fig:fig_ablation}
            \vspace{-0.5mm}
            \end{figure}

        \begin{figure*}[t]
            \centering
            \includegraphics[width=\textwidth]{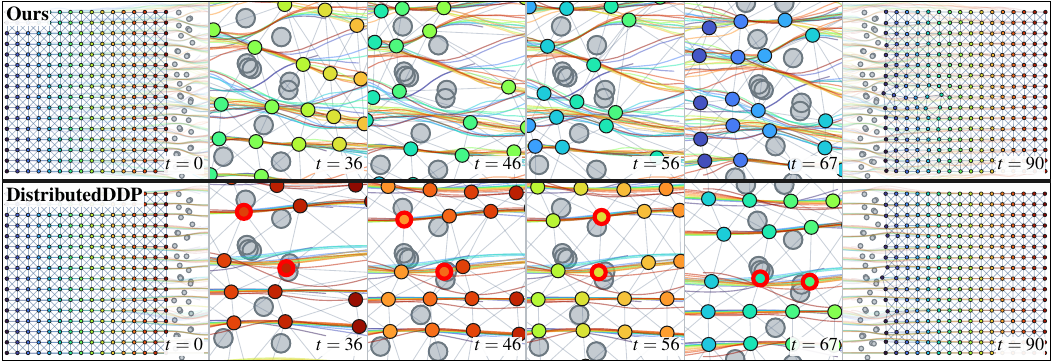}
            \vspace{-7mm}
            \caption{Trajectories in the Formation with 256 agents. Lines between agents indicate local connections. The middle panels magnify instants where a baseline agent penetrates multiple obstacles, with both methods shown over the same spatial and temporal window. Red agent edges indicate collisions.}
        \vspace{-6mm}
        \label{fig:traj_formation}
        \end{figure*}
            
            In this task, we examine interagent coordination with route exploration in cluttered environments. We initialize $N\in\{3,6,12,24\}$ agents evenly on a circle of radius $4\,\mathrm{m}$ with $T=200$. The \textit{antipodal} setting assigns diametrically opposite goals, requiring compatible passing orders near the center, whereas the \textit{permutation} setting randomly assigns goals, requiring agents to navigate through the obstacle field (Fig.~\ref{fig:traj_circle}).
            Each environment contains 11 circular obstacles with $r_{\mathrm{obs}}\sim\mathcal{U}(0.45,0.75)\,\mathrm{m}$ inside the initialization circle with sufficient clearance; obstacles either overlap by at least $0.18\,\mathrm{m}$ or are separated by at least $\frac{8}{3}r_{\mathrm{agent}}$, creating traps while excluding marginally traversable gaps.
            
            Figure~\ref{fig:fig_circle} shows that DS-TO maintains a \textit{Clean} rate of at least $0.98$ for all $N$ under the \textit{permutation} setting, while it decreases to $0.85$ at $N=24$ under the \textit{antipodal} setting. At this scale, both deterministic baselines have zero \textit{Clean} rate under the \textit{antipodal} setting; under the \textit{permutation}, SQP reaches $0.35$ while DistributedDDP reaches zero. As the problem size increases, the gradient-based baselines become increasingly prone to local minima, resulting in lower \textit{Clean} rates.
            DSS has no clean episodes at $N=12$ and exceeds available memory at $N=24$. The computation time for DS-TO remains below $20\,\mathrm{s}$ at $N=24$ in both settings, compared with at least $52.7\,\mathrm{s}$ for DistributedDDP and $632.4\,\mathrm{s}$ for SQP.

            The benefit of parallel Jacobi updates becomes more pronounced in this task. At $N=24$ under the \textit{antipodal} assignment, the QP accounts for $55\%$ of the iteration time, yielding a $1.74\times$ speedup over GS, compared with $1.06$--$1.23\times$ for smaller problems. Jacobi also achieves a higher \textit{Clean} rate, $0.85$ versus $0.66$, indicating that sequential updates do not uniformly improve finite-budget solution quality.

            Figure~\ref{fig:fig_ablation} evaluates the role of the soft penalties. At $N=6$, removing the interagent penalty reduces \textit{Clean} under both assignments, while removing both penalties yields no clean episodes, supporting the use of soft penalties for sampling.

    \subsection{Locally Connected Formation of Large Teams}
    \label{subsec:locally_conneceted}

        We next evaluate scalability under local connectivity. Agents with $r_{\mathrm{agent}}=0.15\,\mathrm{m}$ are arranged on a $0.9\,\mathrm{m}$ grid with formation constraints preserving the initial coordinate-wise ordering, $h_{ij,t}^c=-s_{ij}^c(p_{j,t}^c-p_{i,t}^c)\leq0$, where $s_{ij}^c=\mathrm{sign}(p_{j,0}^c-p_{i,0}^c)$ and $c\in\{x,y\}$. The constraint uses $x$, $y$, or both coordinates for horizontal, vertical, or diagonal neighbors. Each of the 20 environments contains 48 obstacles with $r_{\mathrm{obs}}=0.2\,\mathrm{m}$, sampled in a $2\times18\,\mathrm{m}$ region under the same overlap and gap conditions as in the Circle task. Figure~\ref{fig:traj_formation} shows trajectories at the largest scale.

        We report the agent-wise success rate \textit{Succ.}, defined analogously to \textit{Clean} with the formation constraints, together with $\delta_{\max}$. The horizons are $T=40,40,50,70,90,$ and $90$ for $N=2,8,32,64,128,$ and $256$, respectively. As SQP scales poorly with the number of agents, we omit its results here.
        
        \begin{figure}[h]
        \vspace{-1.5mm}
        \includegraphics[width=0.48\textwidth]{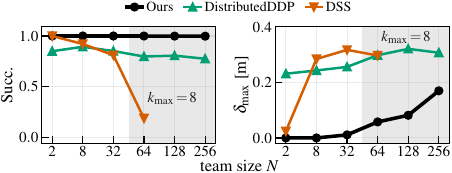}
        \vspace{-2mm}
        \caption{Comparison in the Formation. Left: \textit{Succ.} rate~($\uparrow$); right: maximum penetration~$\delta_{\max}$~($\downarrow$). The shaded region fixes $k_{\max}=8$.}
        \label{fig:fig_formation}
        \vspace{-3mm}
        \end{figure}

            As shown in Fig.~\ref{fig:fig_formation}, DS-TO maintains a success rate of $1.00$ as local connectivity increases over $k_{\max}=1,2,5,$ and $8$, corresponding to $N=2,8,32,$ and $64$. At $k_{\max}=8$, DistributedDDP drops to $0.80$ with $\delta_{\max}=0.298$. DSS also degrades with increasing connectivity; at $N=64$, we reduce its chunk size from $32$ to $16$ to avoid out-of-memory, yet its success rate is only $0.19$.


            Next, fixing $k_{\max}=8$, we scale the team from $N=64$ to $256$. DS-TO maintains a success rate of $1.00$, while DistributedDDP remains between $0.78$ and $0.81$. At $N=256$, DS-TO achieves $\delta_{\max}=0.170$, compared with $0.307$ for DistributedDDP. The primal soft penalties discourage violations before full consensus, whereas consensus residuals in DistributedDDP can directly induce constraint violations.

        \begin{figure}[b]
        \vspace{-4mm}
        \includegraphics[width=0.48\textwidth]{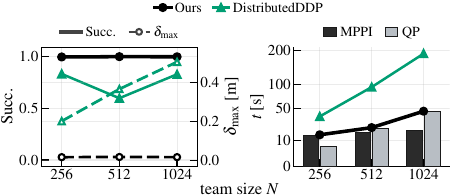}
        \vspace{-2mm}
        \caption{Comparison in the Drone. Left: \textit{Succ.} rate~($\uparrow$) and $\delta_{\max}$~($\downarrow$). Right: average computation time~($\downarrow$) on a square-root scale per framework chunk (in lines), and DS-TO sub-block solves (MPPI and QP; in bars).}
        \label{fig:fig_drone}
        \vspace{-0.5mm}
        \end{figure}
        
\subsection{Scalability in a 3D Multi-Drone Task}
\label{subsec:locally_conneceted_drone}

            We further evaluate scalability on a 3D multi-drone task with $T=90$ over 5 environments. Extending the 2D formation along $z$ gives $k_{\max}=8,17,$ and $26$ for $N=256,512,$ and $1024$. Each environment contains 16 spherical obstacles with $r_{\mathrm{obs}}=0.6\,\mathrm{m}$, using the same $x$--$y$ distribution as in the Formation task and $c_z\sim\mathcal{U}(-a_z,a_z)$, with $a_z=1.65,2.10,$ and $3.00$ chosen according to the formation height to maintain comparable difficulty. The same overlap and gap conditions as in the Circle task are applied in 3D. Due to memory limitations and poor scalability with neighborhood size, we omit DSS from this experiment.
        
            Figure~\ref{fig:fig_drone} shows that DS-TO maintains a success rate of at least $0.997$ with $\delta_{\max}\leq0.017$ across all scales, while DistributedDDP reaches $\delta_{\max}=0.505$ at $N=1024$. At $N=512$, comparable MPPI and QP solve times maximize the benefit of parallel execution, making DS-TO $4.21\times$ faster than DistributedDDP ($22.1\,\mathrm{s}$ compared with $93.0\,\mathrm{s}$). At $N=1024$, the QP dominates as $k_{\max}$ grows, but DS-TO remains $4.19\times$ faster ($44.9\,\mathrm{s}$ compared with $188.3\,\mathrm{s}$).


\section{Conclusion}

We proposed DS-TO, a highly parallel and fully decentralized framework for multi-agent trajectory optimization. DS-TO combines sampling-based optimization in each agent's control space with locally optimized views of interacting agents' trajectories, which are reconciled through decentralized consensus of ADMM. This separates constraint-aware exploration from multi-agent coordination without requiring a centralized optimizer. Across the considered tasks, DS-TO maintained high solution feasibility in challenging non-convex multi-agent problems and scaled to large locally connected teams, including a team of $1024$ drones.

While this work focuses on MPPI as the local solver, the framework remains well suited to other sampling-based methods. Future work will explore integrating more advanced training-free \cite{pan2024modelbased} or learning-based diffusion/flow-based samplers \cite{saravanos2026learningtooptimizeflows} to further enhance performance. Another interesting direction is extending to the MPC setting, where previous solutions would effectively warm-start sampling.



\bibliographystyle{IEEEtran}
\bibliography{references}

@inproceedings{williams2017information,
  title={Information theoretic MPC for model-based reinforcement learning},
  author={Williams, Grady and Wagener, Nolan and Goldfain, Brian and Drews, Paul and Rehg, James M and Boots, Byron and Theodorou, Evangelos A},
  booktitle={2017 IEEE international conference on robotics and automation (ICRA)},
  pages={1714--1721},
  year={2017},
  organization={IEEE}
}

@article{osqp,
  author  = {Stellato, B. and Banjac, G. and Goulart, P. and Bemporad, A. and Boyd, S.},
  title   = {{OSQP}: an operator splitting solver for quadratic programs},
  journal = {Mathematical Programming Computation},
  year    = {2020},
  volume  = {12},
  number  = {4},
  pages   = {637--672},
  doi     = {10.1007/s12532-020-00179-2},
  url     = {https://doi.org/10.1007/s12532-020-00179-2},
}

@book{nocedal2006numerical,
  title={Numerical optimization},
  author={Nocedal, Jorge and Wright, Stephen J},
  year={2006},
  publisher={Springer}
}

@article{boyd2011distributed,
  title={Distributed optimization and statistical learning via the alternating direction method of multipliers},
  author={Boyd, Stephen and Parikh, Neal and Chu, Eric and Peleato, Borja and Eckstein, Jonathan},
  journal={Foundations and Trends{\textregistered} in Machine learning},
  volume={3},
  number={1},
  pages={1--122},
  year={2011},
  publisher={Emerald Publishing Limited}
}

@inproceedings{zhou2021egoswarm,
  author    = {Zhou, Xin and Zhu, Jiangchao and Zhou, Hongyu and Xu, Chao and Gao, Fei},
  title     = {{EGO-Swarm}: A Fully Autonomous and Decentralized Quadrotor Swarm System in Cluttered Environments},
  booktitle = {2021 IEEE International Conference on Robotics and Automation (ICRA)},
  year      = {2021},
  pages     = {4101--4107},
  doi       = {10.1109/ICRA48506.2021.9561902}
}

@article{ferranti2023distributed,
  author  = {Ferranti, Laura and Lyons, Lorenzo and Negenborn, Rudy R. and Keviczky, Tam{\'a}s and Alonso-Mora, Javier},
  title   = {Distributed Nonlinear Trajectory Optimization for Multi-Robot Motion Planning},
  journal = {IEEE Transactions on Control Systems Technology},
  year    = {2023},
  volume  = {31},
  number  = {2},
  pages   = {809--824},
  doi     = {10.1109/TCST.2022.3211130}
}

@article{wurman2008coordinating,
  author  = {Wurman, Peter R. and D'Andrea, Raffaello and Mountz, Mick},
  title   = {Coordinating Hundreds of Cooperative, Autonomous Vehicles in Warehouses},
  journal = {AI Magazine},
  year    = {2008},
  volume  = {29},
  number  = {1},
  pages   = {9--20},
  doi     = {10.1609/aimag.v29i1.2082}
}

@article{saravanos2023distributed,
  author  = {Saravanos, Augustinos D. and Aoyama, Yuichiro and Zhu, Hongchang and Theodorou, Evangelos A.},
  title   = {Distributed Differential Dynamic Programming Architectures for Large-Scale Multiagent Control},
  journal = {IEEE Transactions on Robotics},
  year    = {2023},
  volume  = {39},
  number  = {6},
  pages   = {4387--4407},
  doi     = {10.1109/TRO.2023.3319894}
}

@INPROCEEDINGS{saravanos2024dmpcs,
  author={Saravanos, Augustinos D. and Balci, Isin M. and Bakolas, Efstathios and Theodorou, Evangelos A.},
  booktitle={2024 IEEE/RSJ International Conference on Intelligent Robots and Systems (IROS)}, 
  title={Distributed Model Predictive Covariance Steering}, 
  year={2024},
  volume={},
  number={},
  pages={5740-5747},
  doi={10.1109/IROS58592.2024.10802104}}

@inproceedings{sharon2012conflict,
  author    = {Sharon, Guni and Stern, Roni and Felner, Ariel and Sturtevant, Nathan},
  title     = {Conflict-Based Search for Optimal Multi-Agent Path Finding},
  booktitle = {Proceedings of the AAAI Conference on Artificial Intelligence},
  year      = {2012},
  volume    = {26},
  number    = {1},
  pages     = {563--569},
  doi       = {10.1609/aaai.v26i1.8140}
}

@article{saravanos2026distributedcs,
  title={Distributed Covariance Steering via Non-Convex ADMM for Large-Scale Multi-Agent Systems},
  author={Saravanos, Augustinos D and Balci, Isin M and Abdul, Arshiya Taj and Bakolas, Efstathios and Theodorou, Evangelos A},
  journal={arXiv preprint arXiv:2604.04499},
  year={2026}
}

@article{wang2017safety,
  author  = {Wang, Li and Ames, Aaron D. and Egerstedt, Magnus},
  title   = {Safety Barrier Certificates for Collisions-Free Multirobot Systems},
  journal = {IEEE Transactions on Robotics},
  year    = {2017},
  volume  = {33},
  number  = {3},
  pages   = {661--674},
  doi     = {10.1109/TRO.2017.2659727}
}

@inproceedings{chen2017decentralized,
  author    = {Chen, Yu Fan and Liu, Miao and Everett, Michael and How, Jonathan P.},
  title     = {Decentralized Non-Communicating Multiagent Collision Avoidance with Deep Reinforcement Learning},
  booktitle = {2017 IEEE International Conference on Robotics and Automation (ICRA)},
  year      = {2017},
  pages     = {285--292},
  doi       = {10.1109/ICRA.2017.7989037}
}

@inproceedings{lecleach2020algames,
  author    = {Le Cleac'h, Simon and Schwager, Mac and Manchester, Zachary},
  title     = {{ALGAMES}: A Fast Solver for Constrained Dynamic Games},
  booktitle = {Proceedings of Robotics: Science and Systems},
  year      = {2020},
  doi       = {10.15607/RSS.2020.XVI.091}
}

@article{chen2023multiagent,
  author  = {Chen, Ruishuang and Liang, Zhihui and Cheng, Jie and You, Pengcheng and Yang, Zaiyue},
  title   = {Multi-Agent Cooperative Motion Planning Based on Alternating Direction Method of Multipliers},
  journal = {IEEE Control Systems Letters},
  year    = {2023},
  volume  = {7},
  pages   = {3307--3312},
  doi     = {10.1109/LCSYS.2023.3324663}
}

@article{jiang2024distributed,
  author  = {Jiang, Chao},
  title   = {Distributed Sampling-Based Model Predictive Control via Belief Propagation for Multi-Robot Formation Navigation},
  journal = {IEEE Robotics and Automation Letters},
  year    = {2024},
  volume  = {9},
  number  = {4},
  pages   = {3467--3474},
  doi     = {10.1109/LRA.2024.3368794}
}

@article{wang2022sampling,
  title={Sampling-based optimization for multi-agent model predictive control},
  author={Wang, Ziyi and Saravanos, Augustinos D and Almubarak, Hassan and So, Oswin and Theodorou, Evangelos A},
  journal={arXiv preprint arXiv:2211.11878},
  year={2022}
}

@article{dergachev2025corl,
  title={CoRL-MPPI: Enhancing MPPI With Learnable Behaviours For Efficient And Provably-Safe Multi-Robot Collision Avoidance},
  author={Dergachev, Stepan and Pshenitsyn, Artem and Panov, Aleksandr and Skrynnik, Alexey and Yakovlev, Konstantin},
  journal={arXiv preprint arXiv:2511.09331},
  year={2025}
}

@inproceedings{dergachev2025decentralized,
  title={Decentralized uncertainty-aware multi-agent collision avoidance with model predictive path integral},
  author={Dergachev, Stepan and Yakovlev, Konstantin},
  booktitle={2025 IEEE/RSJ International Conference on Intelligent Robots and Systems (IROS)},
  pages={12456--12463},
  year={2025},
  organization={IEEE}
}

@article{yoon2025distributed,
  title={Distributed Stochastic Search for Multi-Agent Model Predictive Control},
  author={Yoon, Taehyun and Saravanos, Augustinos D and Theodorou, Evangelos A},
  journal={arXiv preprint arXiv:2510.18211},
  year={2025}
}

@inproceedings{summers2012distributed,
  title={Distributed model predictive consensus via the alternating direction method of multipliers},
  author={Summers, Tyler H and Lygeros, John},
  booktitle={2012 50th annual allerton conference on communication, control, and computing (Allerton)},
  pages={79--84},
  year={2012},
  organization={IEEE}
}

@article{zhou2026aclm,
  title={ACLM: ADMM-Based Distributed Model Predictive Control for Collaborative Loco-Manipulation},
  author={Zhou, Ziyi and Shu, Pengyuan and Cao, Ruize and Zhao, Yuntian and Zhao, Ye},
  journal={arXiv preprint arXiv:2603.07095},
  year={2026}
}

@inproceedings{shorinwa2020scalable,
  title={Scalable collaborative manipulation with distributed trajectory planning},
  author={Shorinwa, Ola and Schwager, Mac},
  booktitle={2020 IEEE/RSJ International Conference on Intelligent Robots and Systems (IROS)},
  pages={9108--9115},
  year={2020},
  organization={IEEE}
}

@article{kobilarov2012cross,
  author  = {Marin Kobilarov},
  title   = {Cross-Entropy Motion Planning},
  journal = {The International Journal of Robotics Research},
  volume  = {31},
  number  = {7},
  pages   = {855--871},
  year    = {2012},
  doi     = {10.1177/0278364912444543}
}

@article{gandhi2021robust,
  title={Robust model predictive path integral control: Analysis and performance guarantees},
  author={Gandhi, Manan S and Vlahov, Bogdan and Gibson, Jason and Williams, Grady and Theodorou, Evangelos A},
  journal={IEEE Robotics and Automation Letters},
  volume={6},
  number={2},
  pages={1423--1430},
  year={2021},
  publisher={IEEE}
}

@article{yin2023shield,
  title={Shield model predictive path integral: A computationally efficient robust mpc method using control barrier functions},
  author={Yin, Ji and Dawson, Charles and Fan, Chuchu and Tsiotras, Panagiotis},
  journal={IEEE Robotics and Automation Letters},
  volume={8},
  number={11},
  pages={7106--7113},
  year={2023},
  publisher={IEEE}
}

@article{luukkonen2011modelling,
  title={Modelling and control of quadcopter},
  author={Luukkonen, Teppo},
  journal={Independent research project in applied mathematics, Espoo},
  volume={22},
  number={22},
  pages={1--24},
  year={2011}
}

@inproceedings{
saravanos2026learningtooptimizeflows,
title={Learning-to-Optimize via Deep Unfolded Flows},
author={Augustinos D Saravanos and Oswin So and H M Sabbir Ahmad and Chuchu Fan},
booktitle={Forty-third International Conference on Machine Learning},
year={2026}
}

@inproceedings{pan2024modelbased,
  author    = {Chaoyi Pan and Zeji Yi and Guanya Shi and Guannan Qu},
  title     = {Model-Based Diffusion for Trajectory Optimization},
  booktitle = {Advances in Neural Information Processing Systems},
  volume    = {37},
  year      = {2024},
  doi       = {10.52202/079017-1846}
}

\end{document}